\documentclass[10pt,twocolumn,letterpaper]{article}

\usepackage{cvpr}
\usepackage{times}
\usepackage{epsfig}
\usepackage{graphicx}
\usepackage{amsmath}
\usepackage{amssymb}
\usepackage{tabularx}
\usepackage{multirow}
\usepackage{mathtools}
\usepackage{booktabs}
\DeclareMathOperator*{\argmax}{arg\,max}
\DeclareMathOperator*{\softarg}{softarg}

\usepackage[pagebackref=true,breaklinks=true,letterpaper=true,colorlinks,bookmarks=false]{hyperref}

\newcommand*{\affaddr}[1]{#1} % No op here. Customize it for different styles.
\newcommand*{\affmark}[1][*]{\textsuperscript{#1}}

\cvprfinalcopy % *** Uncomment this line for the final submission

\ifcvprfinal\fi
\begin{document}

%%%%%%%%% TITLE
\title{DeepStrip: High Resolution Boundary Refinement}

\author{Peng Zhou\affmark[1]\qquad Brian Price\affmark[2]\qquad Scott Cohen\affmark[2] \qquad Gregg Wilensky\affmark[2]\qquad Larry S. Davis\affmark[1]\\
\affaddr{\affmark[1]University of Maryland, College Park \qquad}
\affaddr{\affmark[2]Adobe Research, San Jose}
}

\maketitle
%\thispagestyle{empty}

%%%%%%%%% ABSTRACT
\begin{abstract}
In this paper, we target refining the boundaries in high resolution images given low resolution masks. For memory and computation efficiency, we propose to convert the regions of interest into strip images and compute a boundary prediction in the strip domain. To detect the target boundary, we present a framework with two prediction layers. First, all potential boundaries are predicted as an initial prediction and then a selection layer is used to pick the target boundary and smooth the result. To encourage accurate prediction, a loss which measures the boundary distance in the strip domain is introduced. In addition, we enforce a matching consistency and C0 continuity regularization to the network to reduce false alarms. Extensive experiments on both public and a newly created high resolution dataset strongly validate our approach.
\end{abstract}

%%%%%%%%% BODY TEXT
\section{Introduction}

\begin{figure}[t]
\begin{center}
\includegraphics[width=0.45\textwidth]{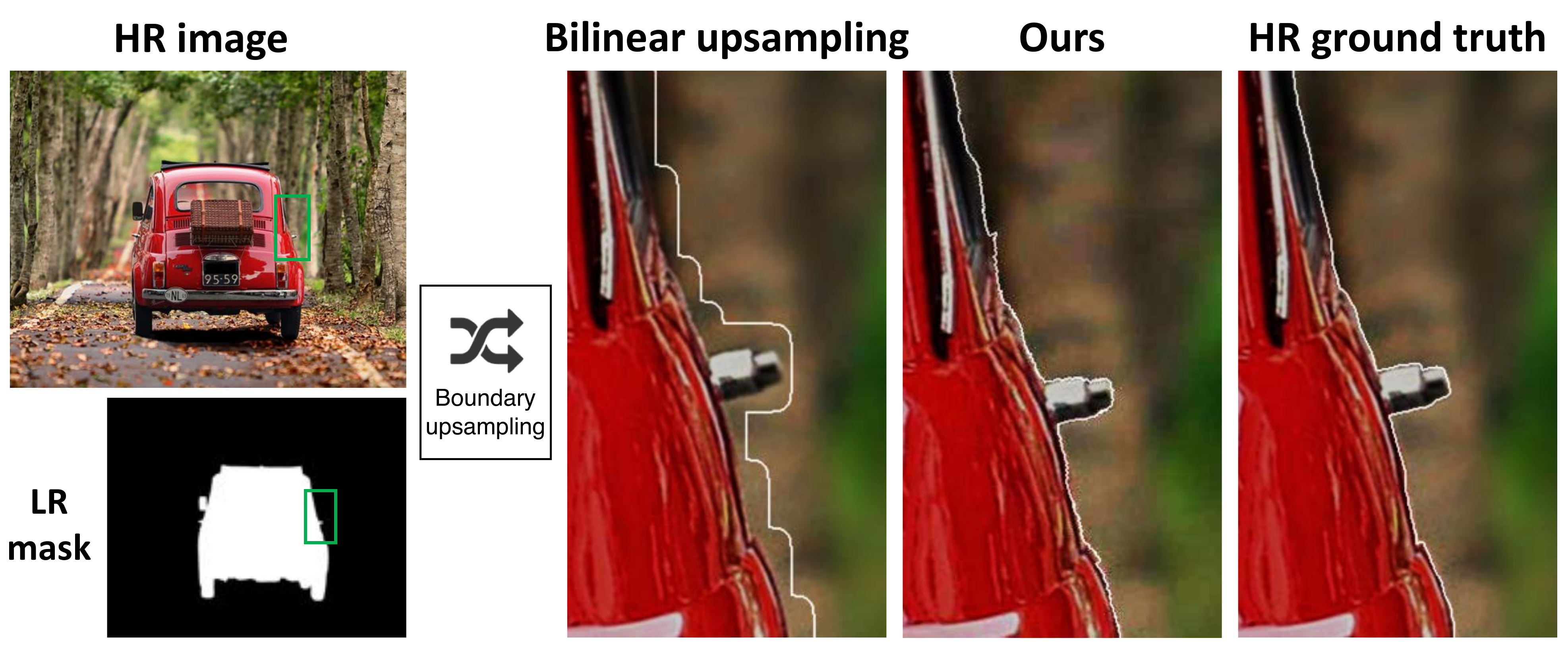}
\end{center}
\vspace{-10pt}
\caption{Concept overview. The example is from the newly created PixaHR dataset. Given low resolution mask and high resolution image on the left, a bilinear upsampling with scale factor $16 \times$ would results in boundary misalignment in high resolution image, as is shown in the enlarged boundary region on the right. Also, the new details in high resolution would be missed.} %(\eg, the new boundary near the car mirror in the rightmost ground truth.)}
\label{fig:concept}
\vspace{-15pt}
\end{figure}

Boundary detection is a well-studied problem and fundamental for human recognition~\cite{opelt2006boundaryfrag,canny1986computational}. Recent decades have witnessed considerable effort to improve the boundary quality of an object that has been detected~\cite{xie15hed,tang2013grabcut,rother2004grabcut,kass1988snakes,wang2018detectglobal,zhao2019pyramid,he2019bi,le2018interactiveboundary} or segmented~\cite{chen2018deeplabv3+,SunXLW19hrnet,li2018interactive,benenson2019large}. Consequently, it is not difficult to separate object of interests from backgrounds with precise boundaries utilizing these methods. While current learning based boundary detection algorithms are usually computed on low resolution (LR) images (0.04-0.25 million pixels), most photos taken these days are much larger, ranging from cell phone size (8-16 million pixels) to professional camera size (16-400 million pixels). Most methods are not designed for images of this size and the excessive computation they require, and most machine learning based methods cannot process them due to memory constraints. Given a precise low resolution prediction, a workaround would be to directly apply upsampling to reach high resolution (HR). Nevertheless, this usually yields poor quality results because the semantic contents in the HR image are not considered. (See Figure~\ref{fig:concept}.)

\begin{figure*}[t]

\begin{center}
\includegraphics[width=1\textwidth]{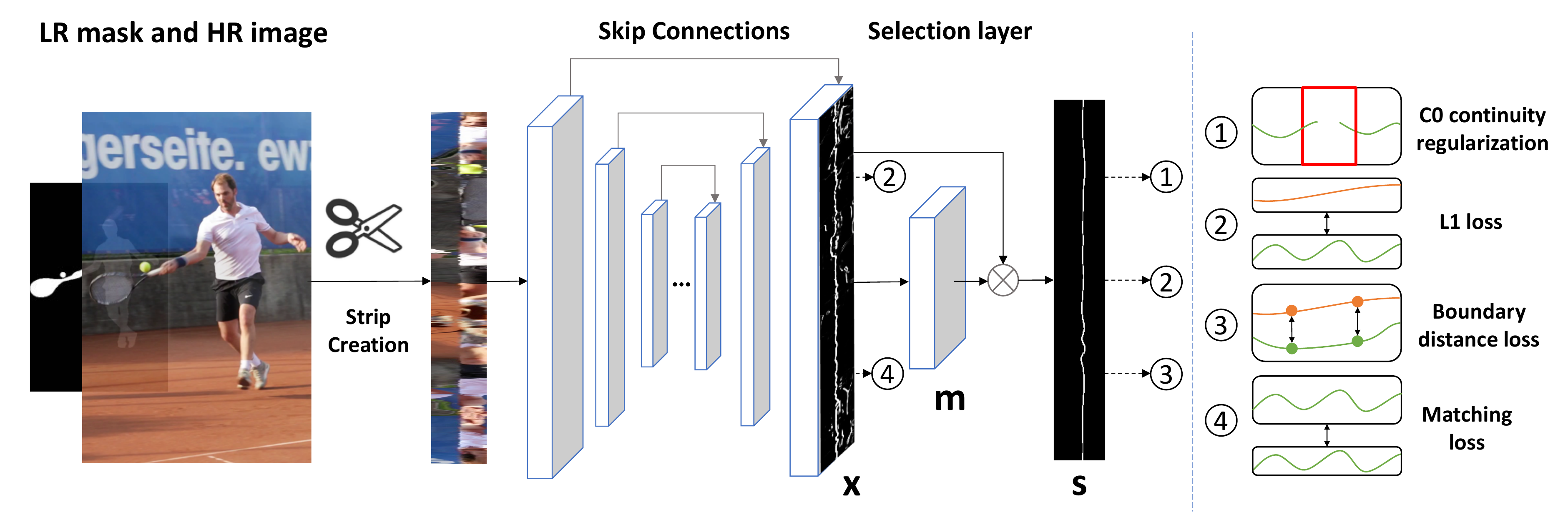}
\end{center}
\vspace{-15pt}
\caption{Framework. To save memory and computation, we predict the boundary in a strip image instead of the whole image. First, the strip image is extracted from the HR image and corresponding LR mask. Feeding the strip image as input, the network predicts all potential boundaries (denoted as ``x") and passes the initial prediction to a selection layer (denoted as ``m") to pursue more accurate prediction on the target boundary (denoted as ``s"). The numbers are indicator to the losses displayed on the right. Orange and green curves denote the ground truth and prediction, respectively. Note that the strip image and prediction are rotated 90 degree for visualization.}
\label{fig:frame}
\vspace{-15pt}
\end{figure*}

Most research in boundary detection focuses on improving the boundary quality in LR through introducing more semantic information~\cite{acuna2019devil,yu2018seal,liu2017richer} or human interaction~\cite{le2018interactiveboundary,CurveGCN2019,xu2016deepgrab,li2018interactive,benenson2019large}. While there has been some work on  HR semantic segmentation~\cite{chen2019GLNET,zhao2018icnet} and upsampling~\cite{wu2018fastguided,zhang2018residualsr}, there is less focus on accurately capturing the boundary detail in HR. Instead of treating this problem as an upsampling problem, we treat it as boundary detection and harness the contents in HR images for prediction.

To this end, we propose a novel approach to handle boundary refinement in HR images. (See Figure~\ref{fig:frame}.) Our key idea is to allow the power of deep learning methods to be applied to HR images in a time and memory efficient manner by operating on narrow images made up of pixels near the boundary. Given an accurate LR mask, the boundary in HR is likely in proximity to the upsampled LR boundary. (See Figure~\ref{fig:concept}.) Therefore, to save memory and computation, we propose to search for the target boundary in a strip region near the boundary of the upsampled mask. The strip image is formed by sampling pixels along and normal to the upsampled mask boundary. Since the normals may not be smooth due to inaccurate boundaries in the upsampled mask, we represent the LR boundary with a spline approximation and directly treat the orthogonal derivatives of the upsampled spline as the normal directions. Feeding as input the generated strip images, we train a network to firstly predict all potential boundaries. Based on the initial prediction, an additional selection layer is included to predict the target boundary more accurately. To encourage closer prediction and reduce false positives, we propose loss functions to minimize the boundary distance between the prediction and ground truth in the strip image and to encourage C0 continuity in the prediction. Lastly, we pursue consistent results through matching the prediction under different strip sizes to further boost the performance.

To validate our approach, we create a new PixaHR dataset (see Figure~\ref{fig:concept} for image example) consisting of 100 photos with average resolution $7k \times 7k$ and evaluate our approach up to scale factor $32 \times$. Results on DAVIS 2016 and COCO coarse annotations also show our ability to refine coarse boundary annotations.

In a nutshell, our contribution is three-fold. 1) We propose an approach to predict the boundary in a strip image which converts potential boundary regions into a strip space. This approach allows us to apply neural networks in a computationally and memory efficient manner. 2) To improve performance and encourage closer prediction, we propose novel losses including boundary distance, matching and C0 continuity loss. 3) We create a high resolution dataset for evaluation. To the best of our knowledge, we are the first learning based approach to make HR dense boundary refinement with resolution up to $10k \times 10k$. Extensive experiments on both public and the new PixaHR dataset strongly highlight our effectiveness.

\section{Related Work}

\noindent\textbf{Boundary Refinement}. 
Multiple attempts have been made to improve boundary quality through extracting better features~\cite{xie15hed,yang2016object,liu2017richer,acuna2019devil,deng2018crisp}. Xie \etal~\cite{xie15hed} utilize features from multiple layers and fuse both low and high level features to detect edges. Liu \etal~\cite{liu2017richer} explore rich convolutional features to boost the performance. More related, attention has been taken to refine coarse boundary predictions or annotations~\cite{yu2018seal,acuna2019devil}. Conventional methods like dense Conditional Random Fields (CRF)~\cite{krahenbuhl2011crf}, Graph Cuts~\cite{boykov2001graphcut} model the relationship between nearby pixels and thus can be applied to refine LR masks~\cite{li2004lazy}. However, these are segmentation based and only low-level features have been utilized. With more supervision, Yu \etal~\cite{yu2018seal} propose to simultaneously learn and align edges to refine misaligned boundaries directly. Acuna \etal~\cite{acuna2019devil} further improve the performance by introducing a thinning layer and active alignment strategy to obtain refined boundary. These methods mainly explore edge detection in LR images. In contrast, we tackle HR boundary refinement and apply detection only on regions around upsampled LR boundary splines and thus is more memory and computation efficient.

\noindent\textbf{Active Contours}. Active contour models like Snakes~\cite{kass1988snakes} have been introduced to refine boundaries from coarse ones. Various approaches have been explored to handle the limitation of Snakes through, \eg, better initialization, morphological operation~\cite{alvarez2010morphological} or user interaction~\cite{le2018interactiveboundary}. Since our method also refines the curve upsampled from LR mask, we can benefit from these methods and refine the boundary further. Instead of taking the whole image as input, deep active contour~\cite{rupprecht2016deepactive} learns to predict the flow of boundary pixels in a patch by patch fashion. However, it cannot guarantee a continuous boundary prediction. Instead, our approach directly extracts a consecutive boundary region and thus contains more global information. Rather than predict the entire curve, other works have explored predicting control points~\cite{castrejon2017polygonrnn,acuna2018polygonrnn++,CurveGCN2019} through recurrent neural networks or Graph Convolutional Networks (GCN)~\cite{kipf2016gcn} and then fit a curve as the final prediction. However, boundary details are smoothed in the spline representation. In contrast, our approach predicts precise edge information directly. Another line of work implicitly represents boundary curves. For example, deep level set methods~\cite{osher1988fronts} evolve boundary curves by minimizing the level energy function. Other learning based approaches~\cite{marcos2018dsac,cheng2019darnet,wang2019delse} have proposed to provide useful features, including texture, color or shape, for better optimization. However, these learning based approaches suffer from computation and memory issues when the resolution increases because they process the entire image while our approach only focuses on the regions around upsampled LR boundaries, and thus requires less computation and memory overhead.  

\noindent\textbf{High Resolution Up-sampling}. With the information of low resolution masks, researchers have focused on achieving high quality HR segmentation masks. Conventional methods~\cite{kopf2007joint, barron2016fastbilateral} reach HR by applying upsampling jointly with the LR mask reference. However, the fixed filter structures have difficulty capturing new HR boundary details. He \etal~\cite{he2012guided} propose guided filtering to smooth while preserving edge information when upsampling. Wu \etal~\cite{wu2018fastguided} make the guided filter faster and learnable. For HR segmentation approaches, Zhao \etal~\cite{zhao2018icnet} propose to aggregate LR features for HR segmentation and Chen \etal~\cite{chen2019GLNET} align both global and local features to avoid heavy GPU consumption for HR segmentation. Even though these methods can be potentially adapted to boundary refinement, our method mainly focuses on boundary regions and is designed to detect boundaries in HR directly. Therefore, our approach learns new HR boundaries better, especially when LR boundaries are coarsely annotated.

\section{Approach}

Our goal lies in refining boundaries in HR images given LR precise masks. To achieve this purpose efficiently, we propose to predict on a strip image that captures the potential boundary region rather than the entire HR image. Figure~\ref{fig:frame} illustrates our framework. Our approach consists of strip image creation, which converts HR RGB image into strip image, strip boundary prediction, which refines the edges on the strip image using a network and strip reconstruction which reconstructs the prediction in the original image from the strip boundary prediction during testing.

\begin{figure}[t]

\begin{center}
\includegraphics[width=0.45\textwidth]{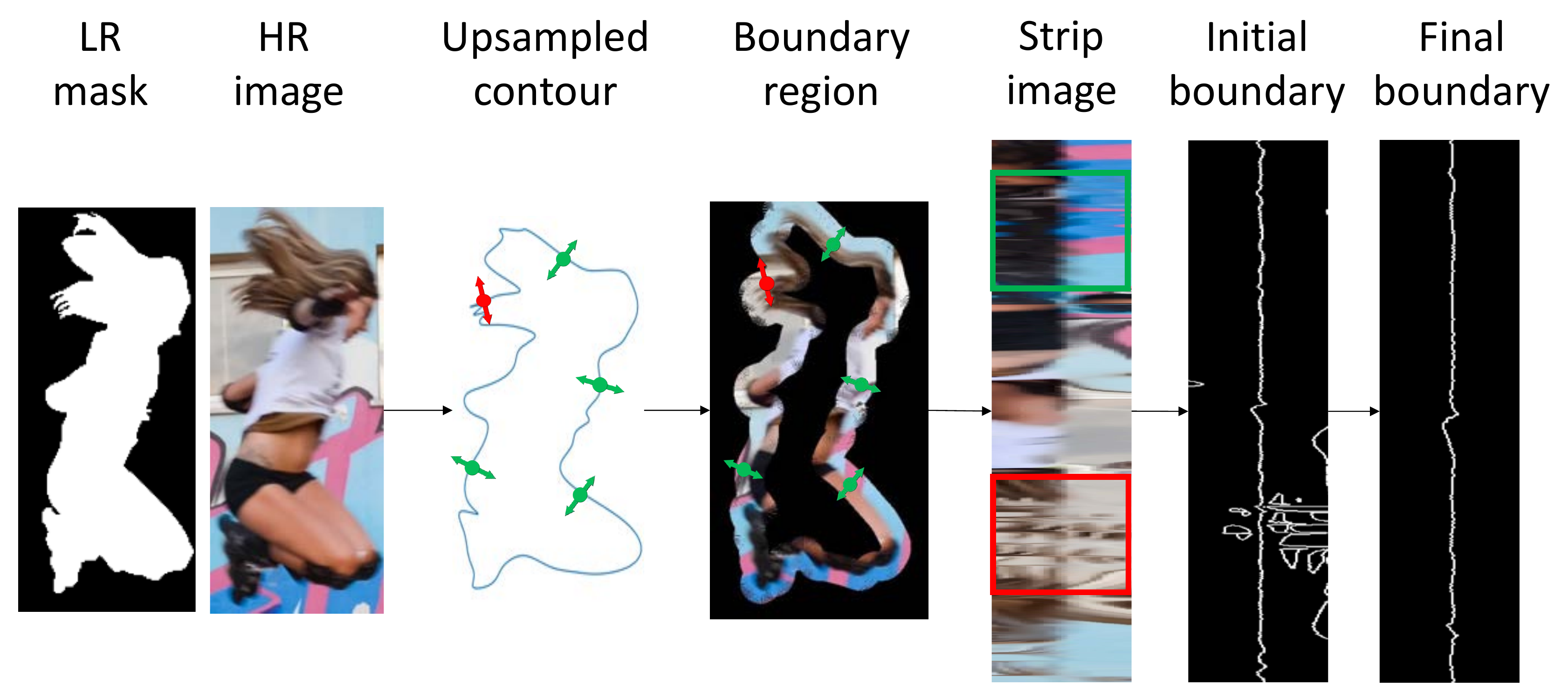}
\end{center}
\vspace{-12pt}
\caption{Strip image creation. To generate strip image, B-spline representation of the contour in the LR mask is upsampled to HR as a coarse boundary. The HR region along the normal direction (\eg, red and green arrows) of the contour is then extracted. Finally, the strip image and corresponding boundary ground truth is obtained by flattening the extracted region in both the HR image and mask. Note that the final boundary filters out noisy boundaries (\eg, the red box region) from the initial boundary. The strip image and boundaries are rotated 90 degree for visualization.}
\label{fig:strip}
\vspace{-12pt}
\end{figure}

\subsection{Strip Image Creation}
Figure~\ref{fig:strip} describes the procedure of strip image creation. Due to the interpolation introduced by upsampling, a directly upsampled boundary from the LR image is likely to be shifted from the ground truth boundary in HR. To localize the real HR boundary pixels, searching around the upsampled boundary is more necessary than searching the whole image. Therefore, we extract pixels near the upsampled boundary to create a strip image. To create the strip image, we step along the boundary and sample points along the normal direction at each point on the curve. To obtain smoothly varying normal directions along the coarse boundary, we represent the LR boundary by B-spline and upsample the LR spline to HR.

Given the HR image $I(p,q)$ and the upsampled spline representation $C=(p(k),q(k))$ of the boundary contour, where $(p(k),q(k))$ denotes the HR image coordinates parameterized by arclength $k$ along the curve, the continuous strip image $J_{I,C}$ is defined by
\begin{equation}
J_{I,C}(k,t\!+\!H/2) \!=\! I(p(k)+t\times\! n_p(k),q(k)+t\times\! n_q(k)),
\end{equation}
where $t$ denotes the distance in the normal direction, $H$ denotes the height of the strip image, and $(n_p(k),n_q(k))$ is the unit normal to the curve at arclength $k$. Accordingly, the strip image $J_{I,C}(j,i)$ with dimension $H \times W$ is obtained by sampling $k=j\times dk$, $t=i\times dt$, where tangential step size $dk=\lfloor |C|/W \rfloor$ and normal step size $dt$ is set to 1 for simplicity. $|C|$ denotes the length of C, $j=0,1,...,W$ and $i=-H/2,...,0,...,H/2$. Also, bilinear interpolation is applied in the high resolution image to evaluate $I(p,q)$ for non-pixel coordinates $(p,q)$.

The corresponding HR strip boundary ground truth is obtained similarly with two adaptations.
First,
for large sampling scale factors, the ground truth boundary is likely to be outside the range of the strip if the strip height is small, making the boundary in strip image not continuous. %(See Figure~\ref{fig:strip})
We add labels at the border of strip if no boundary pixel is included to maintain the C0 continuity of the boundary pixels in the strip image.
Second,
if the strip height is large, multiple boundary pixels might be included in each column in regions where the boundaries are closer than the strip height. In this case, we filter out the extraneous boundaries that are not connected to the current boundary. (See Figure~\ref{fig:strip}.)

\subsection{Strip Boundary Prediction}
Provided the HR strip image as input, we train a network to predict the corresponding boundaries within the strip domain. For memory efficiency, we adapt light-weighted encoder-decoder based structure nested U-Net~\cite{ronneberger2015unet,zhou2018unet++} for boundary prediction. Given the fact that proper dimension of strip image varies for different resolutions, we use instance normalization~\cite{ulyanov2016instance} during training so that the mean and variance are approximated per image. 

As is shown in Figure~\ref{fig:frame}, two prediction layers are proposed to learn the target boundary in strip image to account for the fact that multiple true boundaries may be present in a single column of the strip image. Firstly, we extract the last upsampling layer to predict all potential boundaries. This encourages the network to learn boundary features within the strip image. To predict the target boundary, we add a learnable selection layer to pick up the target boundary from potential boundaries. The input to the selection layer is the initial prediction, and we apply column-wise softmax to the output of the selection layer as a confidence score for the initial prediction. Finally, the target boundary is computed by the multiplication between the initial prediction and the selection score.  The selection layer also smooths the initial prediction, analogous to the non-maximum suppression in Canny edge detection~\cite{canny1986computational}. Formally, 
\begin{equation}\label{selection}
s= x \odot m,
\end{equation}
where $\odot$ denotes pixel-wise multiplication, $s$ denotes the final prediction, $x$ denotes the initial prediction which applies Sigmoid activation to the output of the last upsampling layer and $m$ is the softmax activated output of the selection layer.

\subsection{Loss Function}
Our basic loss function for the initial and final boundary prediction is a weighted $l_1$ loss to differentiate the boundary from non-boundary pixels. Formally,
\begin{equation}
    L_\text{e}=\beta\!\!\!\!\sum_{(i,j)\in Y_+}\!\!\!|y_{ij}-s_{ij}| + (1-\beta)\!\!\!\!\sum_{(i,j)\in Y_-}\!\!\!|y_{ij}-s_{ij}|,
\end{equation}
where $Y_+$ and $Y_-$ denote boundary and non-boundary pixels, respectively. $\beta=|Y_-|/|Y|$ denotes the weight to balance the label and $|Y|$ denotes the total number of pixels in strip mask. $s_{ij}$ denotes the prediction and $y_{ij}$ denotes the binary ground truth at position $(i,j)$ in the strip image. 

In addition, we adapt Dice loss~\cite{wong20183dice} to boundary prediction to encourage intersection between prediction and ground truth:
\begin{equation}
L_\text{dice}=1- \frac{2\sum s_{ij} \times y_{ij} + \epsilon}{\sum s_{ij}+\sum y_{ij} + \epsilon},
\end{equation}
 where $\epsilon$ denotes a small constant to avoid zero division. The loss aims to maximize the intersection over union between the prediction and ground truth.  

\subsubsection{Boundary Distance Loss}
For boundary prediction, a closer prediction to the boundary ground truth is preferred. However, both weighted $l_1$ and dice loss are not sensitive to the distance from prediction to ground truth. Therefore, we introduce a boundary distance loss to measure the average distance between the predicted boundary and the ground truth to encourage closer prediction. Thanks to the strip domain which maps the regions along the normal direction in every column, the boundary distance can be calculated directly through the difference between the prediction and ground truth. Given the prior that only one boundary pixel exists in each column in the final strip mask, the boundary distance at every column can be measured by calculating the argmax difference at every column between the prediction and ground truth. Since argmax function is not differentiable, we approximate it through soft argmax before calculating the boundary distance and formulate the loss as
\begin{equation}
L_\text{d}=\frac{1}{W} \sum^{W}_{j=1}|\displaystyle \softarg_i(s_{ij})-\displaystyle \argmax_i(y_{ij})|,
\end{equation}
where $W$ is the width of strip mask and 
%$\displaystyle \softarg_i(s_{ij})$ is 
the soft argmax in each column (normal direction) is computed as
\begin{equation}
\displaystyle \softarg_i(s_{ij}) = \sum^{H}_{i=1}\left(\frac{|s_{ij}|}{||S_j||_1}\times i\right),
\end{equation}
where $||S_j||_1$ is the $l_1$ normalization of $s_{ij}$ at column $j$. Since the final prediction $s_{ij}$ encourages a unimodal distribution according to Equation~\ref{selection}, this loss enforces the column-wise maximum activation of the final prediction to match with that in ground truth.

\subsubsection{Matching Loss}

Since the strip height is fixed during training, to introduce variance and avoid overfitting on specific strip height, we augment the data through cropping the strip height. Starting from a large height, we crop the strip to a shorter one and make a new prediction. For consistency, the overlapped regions between original and the cropped strip should have the same initial prediction since all potential boundaries are predicted. Formally, we take a $l_1$ loss between the cropped and original initial prediction to calculate the matching loss,
\begin{equation}
L_\text{m}=\frac{1}{|Y_{crop}|}\sum_{(i,j)\in Y_{crop}}|x'_{ij}-x_{ij}|,
\end{equation}
where $Y_{crop}$ is the cropped region of original mask $Y$ and $x'_{ij}$ is the new initial prediction for the cropped strip image. In addition, this loss also helps the network learn to ignore spurious edges detected near the border of the strip.

\subsubsection{C0 Continuity Regularization}
 Additionally, we add a C0 continuity regularization to the final prediction to enforce a continuous prediction. Ideally, at most one boundary pixel is allowed at every column in the final prediction, so the prediction is C0 continuous if the maximum activated position of every column is C0 continuous. Specifically, we compute the soft argmax of every column, calculate a marginal difference between nearby argmax columns and penalize the position within a window size where prediction becomes discontinuous. Formally,
\begin{equation}
L_\text{C0}\!\!=\!\!\frac{1}{W}\!\sum^W_{j=1}\!P(\mbox{max}(0,\!|\displaystyle \softarg_i(s_{ij})\!-\!\displaystyle \softarg_i(s_{i,j+1})|\!-\!v)\!), 
\end{equation}
where $v$ denotes the margin value and $P$ denotes the maxpooling with a fixed kernel size so that all pixels within the range get penalized. $s_{iW+1}$ is replicated by $s_{i1}$ for calculation. This loss serves as a self regularization as no ground truth label is required.

The total loss function is therefore,
\begin{equation}
L_\text{total}= L_e + L_{dice} + \lambda_1 L_d + \lambda_2 L_m + \lambda_3 L_{C0},
\end{equation}
where $\lambda_1, \lambda_2, \lambda_3$ are hyper-parameters to adjust the weight of each loss. $L_e$ is applied to both the initial and final prediction. $L_m$ is only applied to the initial prediction and $L_{dice}, L_d, L_{C0}$ are applied only to the final prediction. With the total loss function, a closer prediction is preferred and the network draws attention to the target boundaries.

\subsection{Strip Reconstruction}

To make a prediction on the HR image, a mapping between the predicted strip boundaries and the full HR mask is required at inference. For every pixel in the strip image, the corresponding coordinates in the HR image are recorded for reconstruction. Given the raw prediction, we optimize the path with a dynamic programming similar to seam carving~\cite{avidan2007seam} and find the path with minimum energy. We minimize the function
\begin{equation}
E_{ij} = -s_{ij} - \frac{|\partial I(i,j)|}{max(|\partial I|)},
\end{equation}
where $|\partial I(i,j)|$ denotes the magnitude of the image gradients at $(i,j)$. The algorithm searches for the energy cost for neighborhood pixels and finds the path with a minimum energy cost, which indicates the boundary path with the highest probability. We then connect the original coordinates of the final path in the full mask to form the full prediction.

At inference, the flexible input dimension of our framework enables different strip sizes for different images. Benefitting from it, we determine the width of strip, which reflects the number of sampling points along the boundary, by multiplying the LR boundary length with the scale factor. We fix the height of strip with the assumption that all target boundaries are involved, and an adaptive height adjustment strategy is also discussed in Section~\ref{discuss}. For objects containing multiple contours due to complex topology, the prediction is made on each contour separately.

\subsection{Implementation Details}
We generate the spline curve efficiently from the binary mask using the scipy function ‘splprep’ after extracting contours. To guarantee a consistent sign for the normals, we extract strip images from closed contours. The starting point of strip is not deterministic so that no bias is introduced in training. The final ground truth strip boundary mask is obtained by taking the gradient of the ground truth segmentation mask after removing any isolated noisy boundaries. Additionally, we randomly add small shifts to the spline representation to introduce position variation of the target boundary in strip image during training. Our framework is implemented in Pytorch. The encoder consists of 4 $3 \times 3$ convolutional layers and the decoder consists of 4 upsampling layers. The selection layer consists of another convolutional layer with $3 \times 3$ kernel size. The activation function is ReLU~\cite{glorot2011relu} for all encoder and decoder layers. We use instance normalization for all normalization layers to enable flexible input size at inference. During training, the input strip dimension is fixed as $80 \times 4096$. We train the network for 70 epochs with batch size 6 on an NVIDIA GeForce TITAN P6000. We use Stochastic Gradient Descent (SGD) as optimizer and the initial learning rate is 0.1. The learning rate decays by a factor of 10 after every 20 epochs. The momentum is set to 0.9 and weight decay is set to 0.0005. $\lambda_1$, $\lambda_2$ and $\lambda_3$ are set to be 0.1, 20 and 1 empirically. We crop strip image by half to obtain $Y_{crop}$ for matching loss and the maxpooling kernel size for C0 continuity regularization is 11. The margin in C0 continuity regularization is set to 1. Horizontal flipping is applied as data augmentation.

\begin{table*}[t]
\centering
\begin{tabular}{p{3.5cm}cccccccc}
\toprule
 \multicolumn{1}{l}{Dataset}
  & \multicolumn{2}{c}{\textbf{DAVIS 2016}~\cite{perazzi2016davis} $4\times$}  & \multicolumn{2}{c}{\textbf{PixaHR $8\times$}} & \multicolumn{2}{c}{\textbf{PixaHR $16\times$}} &\multicolumn{2}{c}{\textbf{PixaHR $32\times$}}\\ \midrule
 Metrics & {$F$(0 pix)} & {$F$(1 pix)}& {$F$(1 pix)} & {$F$(2 pix)}& {$F$(1 pix)} & {$F$(2 pix)}& {$F$(1 pix)} & {$F$(2 pix)}\\ \midrule
Bilinear Upsampling  &0.171 &0.521 &0.116 &0.194 &0.15&0.187 &0.07& 0.106\\
Grabcut~\cite{rother2004grabcut}   &0.232 & 0.541 &0.063 &0.121 & 0.020 &0.053 &0.0 & 0.0\\
Dense CRF~\cite{krahenbuhl2011crf} &0.268 & 0.702 & 0.278 &0.434 &0.245 &0.389 &0.142 & 0.227 \\ 
Bilateral Solver~\cite{barron2016fastbilateral} &0.274 & 0.569  &0.207   &0.277  &0.185 &0.247  & 0.156 &0.216 \\
Curve-GCN~\cite{CurveGCN2019} &0.076 &0.160 &0.021 &0.033 &0.018 &0.028 &0.012 &0.028 \\
DELSE~\cite{wang2019delse} &0.271 &0.531 &0.096 &0.133 &0.086 &0.132 &0.080 &0.130\\
STEAL~\cite{acuna2019devil} &0.171 &0.348 & 0.282 &0.457 &0.151 &0.255 &0.09 & 0.144 \\
JBU~\cite{kopf2007joint}  &0.175 &0.447 & 0.140 & 0.231 & 0.117 & 0.184  &0.055 &0.090 \\
Guided Filtering~\cite{he2012guided} &0.129 &0.349 &0.121 &0.195 &0.092 &0.145 &0.060 &0.097 \\
Deep GF~\cite{wu2018fastguided}  &0.193 &0.461 & 0.286 &0.420  & 0.175 & 0.269 &0.09 &0.141 \\
U-Net boundary  &0.320 &0.656 & 0.170 &0.297  & 0.139 &0.197 &0.068 &0.108 \\\midrule
U-Net strip (baseline) &0.303 &0.710 &0.334 &0.455  &0.303 &0.425 &0.267 &0.357 \\
Ours & \textbf{0.423}   & \textbf{0.788} &\textbf{0.416}  &\textbf{0.508} &\textbf{0.396}   &    \textbf{0.498} & \textbf{0.330} &\textbf{0.447}\\ \bottomrule 
\end{tabular}
\caption{Boundary-based $F$ score comparison. The scale factor between low and high resolution image is 4 on DAVIS 2016 and 8, 16, 32 on PixaHR. For DAVIS 2016, the pixel dilation is 0 and 1 and for PixaHR is 1 and 2 instead.}
\label{tab:main}
\vspace{-15pt}
\end{table*}

\section{Experiments}\label{exp}
We evaluate our approach on two HR datasets which provide both low and high resolution ground truth in Section~\ref{main}, and then analyze the importance of each components in our framework in Section~\ref{abl}. We also provide memory and speed comparison in Section~\ref{mem}.

\subsection{Datasets and Metrics}
For our experiments, we need a dataset with highly accurate pixel-level HR annotation. Unfortunately, most current datasets are low resolution and many provide inaccurate polygon boundaries as ground truth annotations. We found DAVIS~\cite{perazzi2016davis} to provide accurate enough results with a resolution that is usable for our needs. To better evaluate the results at large scaling factors, we introduce a new dataset---PixaHR. We describe these datasets below.

\noindent\textbf{DAVIS 2016}~\cite{perazzi2016davis}: A benchmark for video segmentation which consists of 50 classes with precise annotations in both 480P and 1080P. To enlarge the scale factor, we down sample the 480P mask by a factor of 2, train our approach on the 30-class 1080P training set with 240P LR masks and test on 20-class 1080P testing set. The scale factor is 4.5 for this experiment. The results are evaluated frame by frame.

\noindent\textbf{PixaHR}: To evaluate more realistic scenarios, we create a PixaHR dataset. It contains 100 images with average resolution $7k \times 7k $ (ranging from $5k \times 5k$ to $10k \times 10k$) collected from public photograph website Pixabay~\cite{pixabay}. We manually annotate the object boundary in the HR images, downsample the HR mask by $8\times$, $16\times$ and $32\times$ and obtain binary LR mask for evaluation. The photos were uploaded by public users and have diverse contents. We apply our model that was trained on DAVIS to this dataset for evaluation.

\noindent\textbf{Metrics}: We use boundary-based F score introduced by Perazzi \etal ~\cite{perazzi2016davis} for evaluation, which is designed to evaluate the boundary quality of segmentation. As it allows changing pixel tolerance by dilation, we set 0 and 1 pixel dilation on DAVIS, and 1 and 2 pixel on PixaHR dataset to measure how close the prediction is to the ground truth. 

\subsection{Main Results}\label{main}

For upsampling based approaches, we compare our approach with \textbf{Bilinear Upsampling}, \textbf{Bilateral Solver~\cite{barron2016fastbilateral}}, Joint Bilateral Upsampling~\cite{kopf2007joint} (\textbf{JBU}), \textbf{Guided Filtering~\cite{he2012guided}} and \textbf{Deep GF~\cite{wu2018fastguided}}. The boundary is obtained by taking the gradient of the upsampled mask. For boundary refinement approaches, we compare with \textbf{Grabcut~\cite{rother2004grabcut}}, \textbf{Dense CRF~\cite{krahenbuhl2011crf}} and \textbf{STEAL~\cite{acuna2019devil}} using upsampled mask as initialization. For active contour methods, the baselines are \textbf{Curve-GCN~\cite{CurveGCN2019}} and \textbf{DELSE~\cite{wang2019delse}}, and predictions on PixaHR are made in LR and upsampled to original resolution since the whole boundary region is required at inference. Learning based approaches are trained or fine-tuned on the training set of DAVIS and evaluated directly on all datasets. More details about baselines are provided in \textbf{supplementary material}. In addition, we also compare our own implemented baselines as below:

\noindent$\bullet$\enskip\textbf{U-Net boundary}: We train U-Net directly on the full resolution images on DAVIS for boundary prediction. We concatenate both the full resolution image and upsampled masks as input so that the network learns to refine the coarse masks. The loss function is a weighted binary cross entropy following Xie \etal~\cite{xie15hed}. Similarly, we also add deep supervision and fuse all intermediate layers to obtain the final prediction. The prediction is made patch-by-patch with patch size $1920 \times 1080$ on PixaHR dataset.

\noindent$\bullet$\enskip\textbf{U-Net strip (baseline)}: Our baseline method which learns to directly predict the target boundary on strip image. Only weighted $l1$ loss is used as loss function.

\noindent$\bullet$\enskip\textbf{Ours}: Our full model which applies selection layer to predict the boundary in strip images with our boundary distance loss, matching loss and C0 continuity regularization.

Table~\ref{tab:main} exhibits our advantage over the baselines. For the DAVIS dataset, a simple upsampling yields a boundary shift from the ground truth and thus performs poorly. Grabcut and dense CRF are segmentation based and thus yield worse performance than ours. Even though other methods including bilateral solver, JBU and Deep GF leverage the low resolution mask, they are designed for general upsampling instead of for boundary refinement and prediction. Curve-GCN fits the curve from the predicted control points which cannot generate as precise a boundary as ours. DELSE moves the contour along the gradient of its energy function, but is less robust than our approach which predicts the target boundary pixels. Additionally, our approach outperforms STEAL as the scale factor increases, indicating the active alignment in STEAL may not be accurate enough for pixel-level boundary prediction. Compared with U-Net boundary, predicting the boundary in strip image (U-Net strip) yields a slightly better performance, perhaps because the strip image narrows down the search space for target boundary. As expected, with our selection layer and proposed losses, we boost the performance further by better determining the target boundaries from other potential boundaries. A similar tendency is observed on PixaHR dataset. Note that in large scale factor 32, most of the methods fail to make close predictions to the ground truth while our method still has a relatively stable performance. 

\begin{table}[t]
\centering  
\small
\begin{tabular}{p{4cm}cc} 
\toprule
 Dataset & DAVIS 2016 & PixaHR $16 \times$\\ \midrule    
 Metrics & $F$(0 pix) & $F$(1 pix)\\ \midrule   
U-Net strip &0.303 & 0.303 \\
U-Net strip dice &0.323 &0.320   \\
U-Net strip dice  \!+ \! selection &0.372 &0.328  \\
U-Net strip dice \!+\! selection \!+\! BD  &0.390 &0.342  \\
Our w/o matching &0.405 &0.365   \\
Ours &0.423 &0.396\\\bottomrule 
\end{tabular}
\caption{Ablation analysis on two datasets. Each entry is the boundary-based F score tested on individual dataset.}
\label{tab:ablation}
\vspace{-5pt}
\end{table}

\begin{table}[t]
\centering  
\small
\begin{tabular}{p{3cm}ccc} 
\toprule
 Methods & Memory (MB) & Speed (s/image) \\ \midrule    
Bilinear Upsampling &- & 0.01/0.02\\
Grabcut~\cite{rother2004grabcut} &- & 5.17/320 \\
Dense CRF~\cite{krahenbuhl2011crf}  & - & 3.22/310\\
Bilateral Solver~\cite{barron2016fastbilateral} &-  &4.18/158 \\
JBU~\cite{kopf2007joint} & -& 0.08/5.71\\
Guided filtering~\cite{he2012guided} &- &0.08/16.1\\
Deep GF~\cite{wu2018fastguided} & - & 0.07/3.95\\
STEAL~\cite{acuna2019devil} &7775/7959 &43.1/4231 \\
Curve-GCN~\cite{CurveGCN2019}  &17330/17330 &  0.93/75.2  \\
DELSE~\cite{wang2019delse}   &17771/17771 &1.02/20.4 \\
U-net boundary &17000/17000 & 0.31/24.5\\
Ours &3300/3300 & 0.28/2.51\\\bottomrule 
\end{tabular}
\caption{Memory and speed comparison. Each entry is the memory or speed on DAVIS 2016/PixaHR dataset. We only compare the memory usage among learning-based approaches.}
\label{tab:speed}
\vspace{-15pt}
\end{table}

\begin{figure*}[t]

\begin{center}
\includegraphics[width=0.9\textwidth]{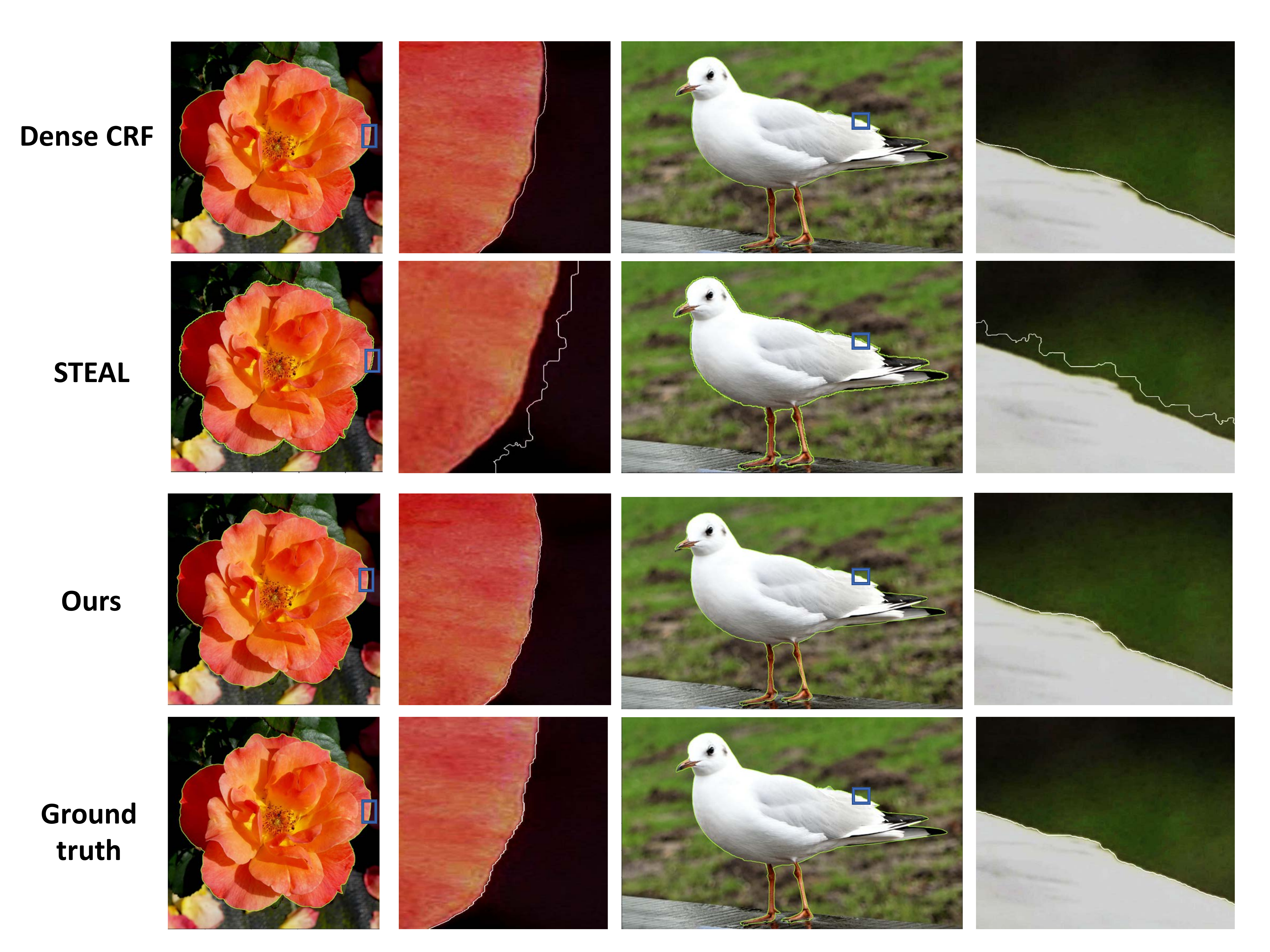}
\end{center}
\vspace{-15pt}
\caption{Qualitative results on PixaHR $32 \times$. Rows from top to down are the results of Dense CRF, STEAL, Ours and the Ground truth. We show the entire boundary (green color) result first and enlarge the blue bounding box region for comparison (boundaries are whitened).}
\label{fig:pixa_vis}
\vspace{-15pt}
\end{figure*}

\begin{figure}[t]

\begin{center}
\includegraphics[width=0.5\textwidth]{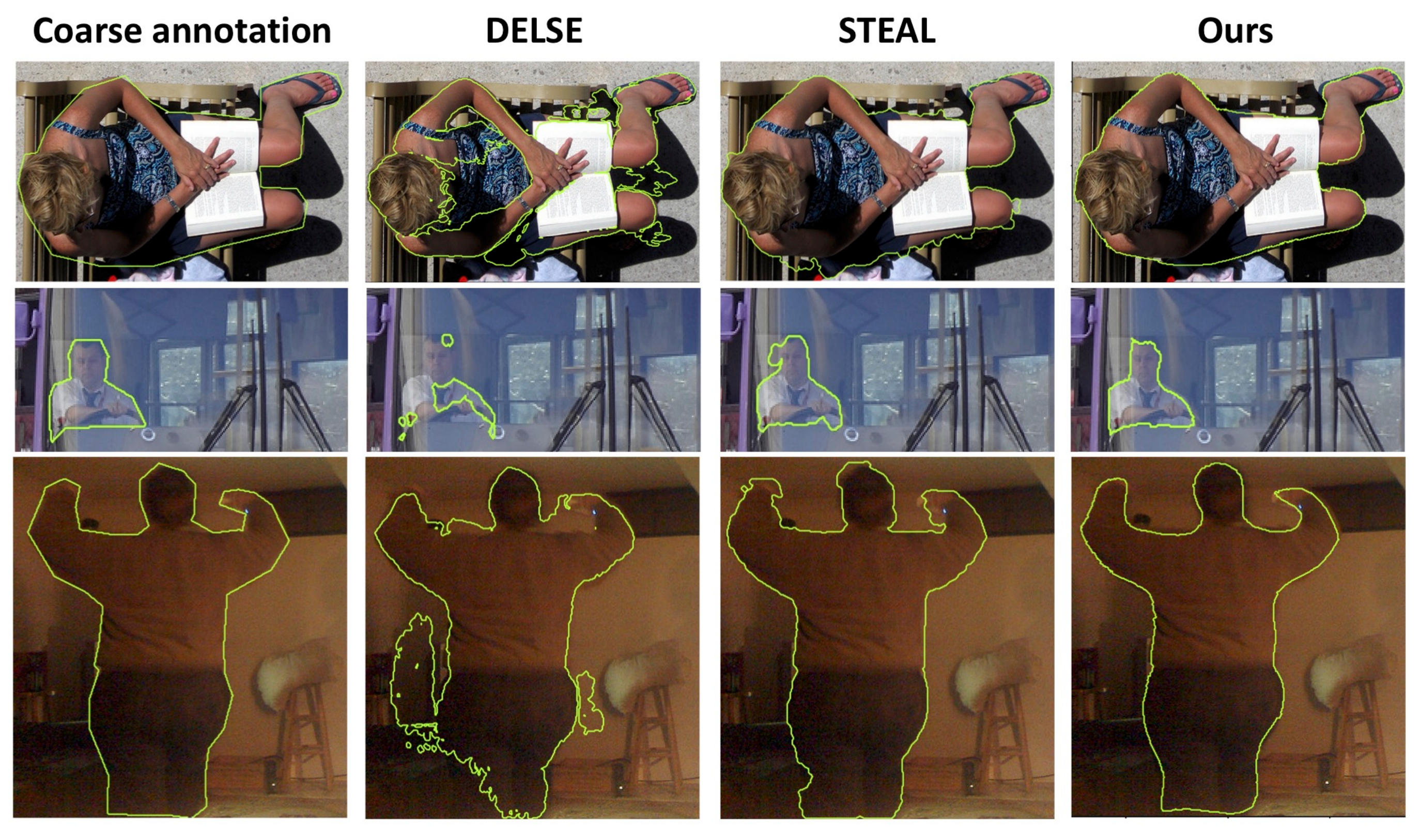}
\end{center}
\vspace{-15pt}
\caption{Qualitative results on COCO. Columns from left to right are coarse annotation, DELSE~\cite{wang2019delse}, STEAL~\cite{acuna2019devil} and Ours. }
\label{fig:qr}
\vspace{-15pt}
\end{figure}

\subsection{Ablation Analysis}\label{abl}
We analyze the importance of each component in our framework as listed below:

\noindent$\bullet$\enskip\textbf{U-Net strip dice}: Adding dice loss to the baseline.

\noindent$\bullet$\enskip\textbf{U-Net strip dice + selection}: Adding dice loss and selection layer to the baseline.

\noindent$\bullet$\enskip\textbf{U-Net strip dice + selection + BD}: Adding dice, boundary distance loss and selection layer to the baseline.

\noindent$\bullet$\enskip\textbf{Ours w/o matching}: Adding additional C0 regularization. It is our full model without the matching loss.

Table~\ref{tab:ablation} summarizes the comparison result. Starting from our baseline U-Net strip, adding dice loss encourages more intersection with the ground truth boundary and thus yields better performance. Comparing \textbf{U-Net strip + dice} with \textbf{U-Net strip + dice + selection}, the selection layer boosts the performance on DAVIS by a large margin, indicating it effectiveness in suppressing the noisy boundaries and smoothing the final prediction. Also, with the boundary distance loss the network learns to have closer prediction. With C0 regularization (\textbf{Ours w/o matching}), the network filters out false positive boundaries by making a continuous prediction. Finally, the performance further improves with the matching loss because the network makes a consistent prediction over different strip heights to avoid overfitting.

\subsection{Memory and Speed Comparison}\label{mem}
Since we only extract a strip image for prediction, our approach is efficient in both memory and computation. Table~\ref{tab:speed} compares our memory overhead and speed performance with baselines. Over all, our computation and memory requirement is relatively small. Our memory requirement is smaller than other learning based approaches. Note that for U-Net boundary and STEAL, the prediction on PixaHR is made patch-by-patch due to the high resolution. 

More specifically, the main computation in our approach lies in strip reconstruction. \eg, for a $1920 \times 1080$ DAVIS image with around 3200 pixels along the boundary, our strip image creation takes 0.08s, prediction process takes 0.06s and the strip reconstruction takes 0.14s. A similar computation percentage is observed on PixaHR also.

\begin{table}[t]
\centering  
\small
\begin{tabular}{p{4.5cm}ccc} 
\toprule
 Dataset & PixaHR $32 \times$\\ \midrule    
 Metrics  & $F$(1 pix)\\ \midrule   
Ours   & 0.330\\
Ours adaptive 1 segment &0.353\\
Ours adaptive 2 segments &0.365\\\bottomrule 
\end{tabular}
\caption{Strip height selection comparison on PixaHR $32\times$.}
\label{tab:discussion}
\vspace{-15pt}
\end{table}

\subsection{Qualitative Results}
We show visualization comparisons in Figure~\ref{fig:pixa_vis}. It is clear that our approach produces more accurate boundariers than the other methods. To further show the effectiveness of our approach on refining the boundaries given LR or coarse masks, we provide qualitative results on COCO where only polygonal boundary ground truth is provided. We directly extract strip image using the coarse annotation on COCO, and visualize the prediction in Figure~\ref{fig:qr}. Comparing with other approaches, our method provides more accurate boundaries, indicating the potential application of our approach to help refine the coarse boundaries. For more visualization results, please see \textbf{supplementary material}.

\subsection{Strip Height Adaptation}\label{discuss}
We predict the target boundary in the strip image under the assumption that the target boundary exists within the pre-defined height range, however, it might not hold true especially for a large scale factor. While one solution is to pre-define a larger height for strip image creation, we propose to progressively increase the height and regenerate strip image to make new predictions at inference. Specifically, we increase the height of strip image until the summation of the final prediction score decreases. Furthermore, height adjustment is more flexible by dividing the whole contour into several segments and adjusting them independently. The results are shown in Table~\ref{tab:discussion}. The comparison between \textbf{Ours} and \textbf{Ours adaptive 1 segment} indicates the effectiveness to have a flexible height. The performance increases further when dividing the whole contour into 2 segments which allows variable height for different regions.

\section{Conclusion}
In summary, this paper presents a novel strategy to handle HR boundary refinement computationally and memory efficiently given LR precise masks. To save memory, we propose to extract boundary regions along the upsampled boundary spline to form a strip image and make prediction within this strip image. To focus on the target boundaries in strip image, boundary distance, matching loss and C0 continuity regularization have been proposed. Extensive experiments on both public and our newly created dataset demonstrate the effectiveness of the proposed approach. However, the current approach still has difficulty predicting complicated topology and soft boundary regions. A smarter adaptive strip height adjustment for every pixel might be a potential solution, which is left for future research.

\section*{Acknowledgement}
This work was partly funded by Adobe. The authors acknowledge the Maryland Advanced Research Computing Center (MARCC) for providing computing resources.

{\small
\bibliographystyle{ieee_fullname}
\bibliography{main}

\begin{thebibliography}{10}\itemsep=-1pt

\bibitem{pixabay}
Pixabay.
\newblock \url{https://pixabay.com}.

\bibitem{acuna2019devil}
David Acuna, Amlan Kar, and Sanja Fidler.
\newblock Devil is in the edges: Learning semantic boundaries from noisy
  annotations.
\newblock In {\em CVPR}, 2019.

\bibitem{acuna2018polygonrnn++}
David Acuna, Huan Ling, Amlan Kar, and Sanja Fidler.
\newblock Efficient interactive annotation of segmentation datasets with
  polygon-rnn++.
\newblock In {\em CVPR}, 2018.

\bibitem{alvarez2010morphological}
Luis {\'A}lvarez, Luis Baumela, Pedro Henr{\'\i}quez, and Pablo
  M{\'a}rquez-Neila.
\newblock Morphological snakes.
\newblock In {\em CVPR}, 2010.

\bibitem{avidan2007seam}
Shai Avidan and Ariel Shamir.
\newblock Seam carving for content-aware image resizing.
\newblock In {\em TOG}, 2007.

\bibitem{barron2016fastbilateral}
Jonathan~T Barron and Ben Poole.
\newblock The fast bilateral solver.
\newblock In {\em ECCV}, 2016.

\bibitem{benenson2019large}
Rodrigo Benenson, Stefan Popov, and Vittorio Ferrari.
\newblock Large-scale interactive object segmentation with human annotators.
\newblock In {\em CVPR}, 2019.

\bibitem{boykov2001graphcut}
Yuri~Y Boykov and M-P Jolly.
\newblock Interactive graph cuts for optimal boundary \& region segmentation of
  objects in nd images.
\newblock In {\em ICCV}, 2001.

\bibitem{canny1986computational}
John Canny.
\newblock A computational approach to edge detection.
\newblock {\em TPAMI}, 1986.

\bibitem{castrejon2017polygonrnn}
Lluis Castrejon, Kaustav Kundu, Raquel Urtasun, and Sanja Fidler.
\newblock Annotating object instances with a polygon-rnn.
\newblock In {\em CVPR}, 2017.

\bibitem{chen2018deeplabv3+}
Liang-Chieh Chen, Yukun Zhu, George Papandreou, Florian Schroff, and Hartwig
  Adam.
\newblock Encoder-decoder with atrous separable convolution for semantic image
  segmentation.
\newblock In {\em ECCV}, 2018.

\bibitem{chen2019GLNET}
Wuyang Chen, Ziyu Jiang, Zhangyang Wang, Kexin Cui, and Xiaoning Qian.
\newblock Collaborative global-local networks for memory-efﬁcient
  segmentation of ultra-high resolution images.
\newblock In {\em CVPR}, 2019.

\bibitem{cheng2019darnet}
Dominic Cheng, Renjie Liao, Sanja Fidler, and Raquel Urtasun.
\newblock Darnet: Deep active ray network for building segmentation.
\newblock In {\em CVPR}, 2019.

\bibitem{deng2018crisp}
Ruoxi Deng, Chunhua Shen, Shengjun Liu, Huibing Wang, and Xinru Liu.
\newblock Learning to predict crisp boundaries.
\newblock In {\em ECCV}, 2018.

\bibitem{glorot2011relu}
Xavier Glorot, Antoine Bordes, and Yoshua Bengio.
\newblock Deep sparse rectifier neural networks.
\newblock In {\em AISTATS}, 2011.

\bibitem{he2019bi}
Jianzhong He, Shiliang Zhang, Ming Yang, Yanhu Shan, and Tiejun Huang.
\newblock Bi-directional cascade network for perceptual edge detection.
\newblock In {\em CVPR}, 2019.

\bibitem{he2012guided}
Kaiming He, Jian Sun, and Xiaoou Tang.
\newblock Guided image filtering.
\newblock {\em TPAMI}, 2012.

\bibitem{kass1988snakes}
Michael Kass, Andrew Witkin, and Demetri Terzopoulos.
\newblock Snakes: Active contour models.
\newblock {\em IJCV}, 1988.

\bibitem{kipf2016gcn}
Thomas~N Kipf and Max Welling.
\newblock Semi-supervised classification with graph convolutional networks.
\newblock {\em ICLR}, 2017.

\bibitem{kopf2007joint}
Johannes Kopf, Michael~F Cohen, Dani Lischinski, and Matt Uyttendaele.
\newblock Joint bilateral upsampling.
\newblock In {\em ToG}, 2007.

\bibitem{krahenbuhl2011crf}
Philipp Kr{\"a}henb{\"u}hl and Vladlen Koltun.
\newblock Efficient inference in fully connected crfs with gaussian edge
  potentials.
\newblock In {\em NeurIPS}, 2011.

\bibitem{le2018interactiveboundary}
Hoang Le, Long Mai, Brian Price, Scott Cohen, Hailin Jin, and Feng Liu.
\newblock Interactive boundary prediction for object selection.
\newblock In {\em ECCV}, 2018.

\bibitem{li2004lazy}
Yin Li, Jian Sun, Chi-Keung Tang, and Heung-Yeung Shum.
\newblock Lazy snapping.
\newblock {\em ToG}, 2004.

\bibitem{li2018interactive}
Zhuwen Li, Qifeng Chen, and Vladlen Koltun.
\newblock Interactive image segmentation with latent diversity.
\newblock In {\em CVPR}, 2018.

\bibitem{CurveGCN2019}
Huan Ling, Jun Gao, Amlan Kar, Wenzheng Chen, and Sanja Fidler.
\newblock Fast interactive object annotation with curve-gcn.
\newblock In {\em CVPR}, 2019.

\bibitem{liu2017richer}
Yun Liu, Ming-Ming Cheng, Xiaowei Hu, Kai Wang, and Xiang Bai.
\newblock Richer convolutional features for edge detection.
\newblock In {\em CVPR}, 2017.

\bibitem{marcos2018dsac}
Diego Marcos, Devis Tuia, Benjamin Kellenberger, Lisa Zhang, Min Bai, Renjie
  Liao, and Raquel Urtasun.
\newblock Learning deep structured active contours end-to-end.
\newblock In {\em CVPR}, 2018.

\bibitem{opelt2006boundaryfrag}
Andreas Opelt, Axel Pinz, and Andrew Zisserman.
\newblock A boundary-fragment-model for object detection.
\newblock In {\em ECCV}, 2006.

\bibitem{osher1988fronts}
Stanley Osher and James~A Sethian.
\newblock Fronts propagating with curvature-dependent speed: algorithms based
  on hamilton-jacobi formulations.
\newblock {\em JCP}, 1988.

\bibitem{perazzi2016davis}
Federico Perazzi, Jordi Pont-Tuset, Brian McWilliams, Luc Van~Gool, Markus
  Gross, and Alexander Sorkine-Hornung.
\newblock A benchmark dataset and evaluation methodology for video object
  segmentation.
\newblock In {\em CVPR}, 2016.

\bibitem{ronneberger2015unet}
Olaf Ronneberger, Philipp Fischer, and Thomas Brox.
\newblock U-net: Convolutional networks for biomedical image segmentation.
\newblock In {\em MICCAI}, 2015.

\bibitem{rother2004grabcut}
Carsten Rother, Vladimir Kolmogorov, and Andrew Blake.
\newblock Grabcut: Interactive foreground extraction using iterated graph cuts.
\newblock In {\em TOG}, 2004.

\bibitem{rupprecht2016deepactive}
Christian Rupprecht, Elizabeth Huaroc, Maximilian Baust, and Nassir Navab.
\newblock Deep active contours.
\newblock {\em arXiv preprint arXiv:1607.05074}, 2016.

\bibitem{SunXLW19hrnet}
Ke Sun, Bin Xiao, Dong Liu, and Jingdong Wang.
\newblock Deep high-resolution representation learning for human pose
  estimation.
\newblock In {\em CVPR}, 2019.

\bibitem{tang2013grabcut}
Meng Tang, Lena Gorelick, Olga Veksler, and Yuri Boykov.
\newblock Grabcut in one cut.
\newblock In {\em ICCV}, 2013.

\bibitem{ulyanov2016instance}
Dmitry Ulyanov, Andrea Vedaldi, and Victor Lempitsky.
\newblock Instance normalization: The missing ingredient for fast stylization.
\newblock {\em CoRR}, 2016.

\bibitem{wang2018detectglobal}
Tiantian Wang, Lihe Zhang, Shuo Wang, Huchuan Lu, Gang Yang, Xiang Ruan, and
  Ali Borji.
\newblock Detect globally, refine locally: A novel approach to saliency
  detection.
\newblock In {\em CVPR}, 2018.

\bibitem{wang2019delse}
Zian Wang, David Acuna, Huan Ling, Amlan Kar, and Sanja Fidler.
\newblock Object instance annotation with deep extreme level set evolution.
\newblock In {\em CVPR}, 2019.

\bibitem{wong20183dice}
Ken~CL Wong, Mehdi Moradi, Hui Tang, and Tanveer Syeda-Mahmood.
\newblock 3d segmentation with exponential logarithmic loss for highly
  unbalanced object sizes.
\newblock In {\em MICCAI}, 2018.

\bibitem{wu2018fastguided}
Huikai Wu, Shuai Zheng, Junge Zhang, and Kaiqi Huang.
\newblock Fast end-to-end trainable guided filter.
\newblock In {\em CVPR}, 2018.

\bibitem{xie15hed}
Saining Xie and Zhuowen Tu.
\newblock Holistically-nested edge detection.
\newblock In {\em CVPR}, 2015.

\bibitem{xu2016deepgrab}
Ning Xu, Brian Price, Scott Cohen, Jimei Yang, and Thomas~S Huang.
\newblock Deep interactive object selection.
\newblock In {\em CVPR}, 2016.

\bibitem{yang2016object}
Jimei Yang, Brian Price, Scott Cohen, Honglak Lee, and Ming-Hsuan Yang.
\newblock Object contour detection with a fully convolutional encoder-decoder
  network.
\newblock In {\em CVPR}, 2016.

\bibitem{yu2018seal}
Zhiding Yu, Weiyang Liu, Yang Zou, Chen Feng, Srikumar Ramalingam, BVK
  Vijaya~Kumar, and Jan Kautz.
\newblock Simultaneous edge alignment and learning.
\newblock In {\em ECCV}, 2018.

\bibitem{zhang2018residualsr}
Yulun Zhang, Yapeng Tian, Yu Kong, Bineng Zhong, and Yun Fu.
\newblock Residual dense network for image super-resolution.
\newblock In {\em CVPR}, 2018.

\bibitem{zhao2018icnet}
Hengshuang Zhao, Xiaojuan Qi, Xiaoyong Shen, Jianping Shi, and Jiaya Jia.
\newblock Icnet for real-time semantic segmentation on high-resolution images.
\newblock In {\em ECCV}, 2018.

\bibitem{zhao2019pyramid}
Ting Zhao and Xiangqian Wu.
\newblock Pyramid feature attention network for saliency detection.
\newblock In {\em CVPR}, 2019.

\bibitem{zhou2018unet++}
Zongwei Zhou, Md~Mahfuzur~Rahman Siddiquee, Nima Tajbakhsh, and Jianming Liang.
\newblock Unet++: A nested u-net architecture for medical image segmentation.
\newblock In {\em DLMIA}. 2018.

\end{thebibliography}
}

\appendix
\section{Appendix}
\subsection{Baseline Details}
In Section 4 of the paper, we compare our method to prior approaches. Here we give specific details on how we applied each prior work.

\noindent$\bullet$\enskip\textbf{Bilinear Upsampling}: Directly bilinearly upsampling the low resolution mask to high resolution. The hard mask is obtained by the optimal threshold the from soft mask and the boundary is obtained by taking the gradient of the upsampled hard mask. 

\noindent$\bullet$\enskip\textbf{Grabcut~\cite{rother2004grabcut}}: We apply grabcut given the upsampled low resolution, and use the boundary mask for evaluation.

\noindent$\bullet$\enskip\textbf{Dense CRF~\cite{krahenbuhl2011crf}}: A non-learning approach based on conditional random field of the nearby pixels. 
Given the upsampled mask and High Resolution (HR) image, we apply dense CRF to refine the mask. The boundary mask is obtained from the gradient of the predicted mask.

\noindent$\bullet$\enskip\textbf{Bilateral Solver~\cite{barron2016fastbilateral}}: A edge-aware smoothing algorithm with fast and robust optimization. We use the publicly released code for evaluation. We provide the upsampled mask as the reference image. The hard mask is obtained by the optimal threshold from the soft mask.

\noindent$\bullet$\enskip\textbf{JBU~\cite{kopf2007joint}}: A Joint Bilateral Upsampling (JBU) algorithm which upsamples the source image taking into account the reference image jointly. We use the contributed opencv function for evaluation. 
We take the Low Resolution (LR) image as the source image and jointly upsample to obtain the output. The hard mask is obtained by the optimal threshold from the soft mask.

\noindent$\bullet$\enskip\textbf{Deep GF~\cite{wu2018fastguided}}: A learnable guided filtering approach that performs pixel-wise image prediction. 
We use the released code for evaluation~\footnote{\url{https://github.com/wuhuikai/DeepGuidedFilter}} and 
we use radius 1 for testing. 

\noindent$\bullet$\enskip\textbf{Guided Filtering~\cite{he2012guided}}: The original guided filter approach. We use the built-in opencv function for evaluation.

\noindent$\bullet$\enskip\textbf{Curve-GCN~\cite{CurveGCN2019}}: A GCN based approach which aims to predict the control points of the contour and fit curve to obtain the final boundary. Instead of random initialization, we provide the upsampled contours as initialization to train the network. The input size is $512 \times 512$ and HR prediction is made by upsampling from LR prediction as the whole boundary region is required for prediction.

\noindent$\bullet$\enskip\textbf{DELSE~\cite{wang2019delse}}: A level-set based approach with extreme points as initialization. We use the released code for evaluation.
Since a ground truth hi-res mask is not available at inference time, instead of extracting extreme points from ground truth mask, we use the upsampled LR mask to extract extreme points for evaluation. The input dimension is $1024 \times 1024$ and predictions on PixaHR are made in low resolution and upsampled to original resolution. We report the optimal threshold for evaluation. The original DELSE setting with ground truth extreme points is also shown in Table~\ref{tab:additional}.

\noindent$\bullet$\enskip\textbf{STEAL~\cite{acuna2019devil}}: A semantic boundary refinement approach which adds a thinning layer and active alignment to refine boundaries from coarse to fine. We use the public released code and model~\footnote{\url{https://github.com/nv-tlabs/STEAL}} and we follow the default patch-by-patch testing with patch size 512 for evaluation.

\noindent$\bullet$\enskip\textbf{U-Net Boundary}: Since it is difficult to implement in the whole image the boundary distance and C0 continuity loss, which are applied in strip domain, we only apply common edge detection loss as in ~\cite{xie15hed}. As a result, the predicted boundaries are thick with high recall and low precision in Boundary-based F score. (See results in Figure~\ref{fig:pixa_whole} and Figure~\ref{fig:davis_whole})

\subsection{Additional Ablation Analysis}
We provide additional ablation results in Table~\ref{tab:additional}. To determine the width of the strip, besides multiplying the pixel number in LR mask with the scale factor, we slightly increase the width further by a factor from 1 to 2. Comparing among Ours 1, Ours 1.5 and Ours 2, the performance changes by a small margin under different factors. We report Ours 1.5 in the main result as a trade off between performance and computation. Additionally, since our strip reconstruction step uses image gradient as part of energy function, we conduct experiment which only uses image gradient to find the minimum path in the strip reconstruction step. As shown in Table~\ref{tab:additional}, the performance degrades by a large margin if we only use gradient (Strip + gradient) because spurious boundaries will be included, indicating the effectiveness of our learning based approach.
\begin{table}[t]
\centering  
\small
\begin{tabular}{p{3.5cm}cc} 
\toprule
 Dataset & DAVIS 2016 & PixaHR $16 \times$\\ \midrule    
 Metrics & $F$(0 pix) & $F$(1 pix)\\ \midrule   
DELSE original &0.275  & 0.082 \\
Strip + gradient &0.165 &0.295   \\
Our 1 &0.414 &0.392   \\
Our 2 &0.416 &0.415   \\
Ours 1.5 &0.423 &0.396\\\bottomrule 
\end{tabular}
\caption{Ablation analysis on two datasets. Each entry is the boundary-based F score tested on individual dataset.}
\label{tab:additional}
\vspace{-5pt}
\end{table}

\subsection{Qualitative Results of Loss Function}
\begin{figure*}[t]

\begin{center}
\includegraphics[width=\linewidth]{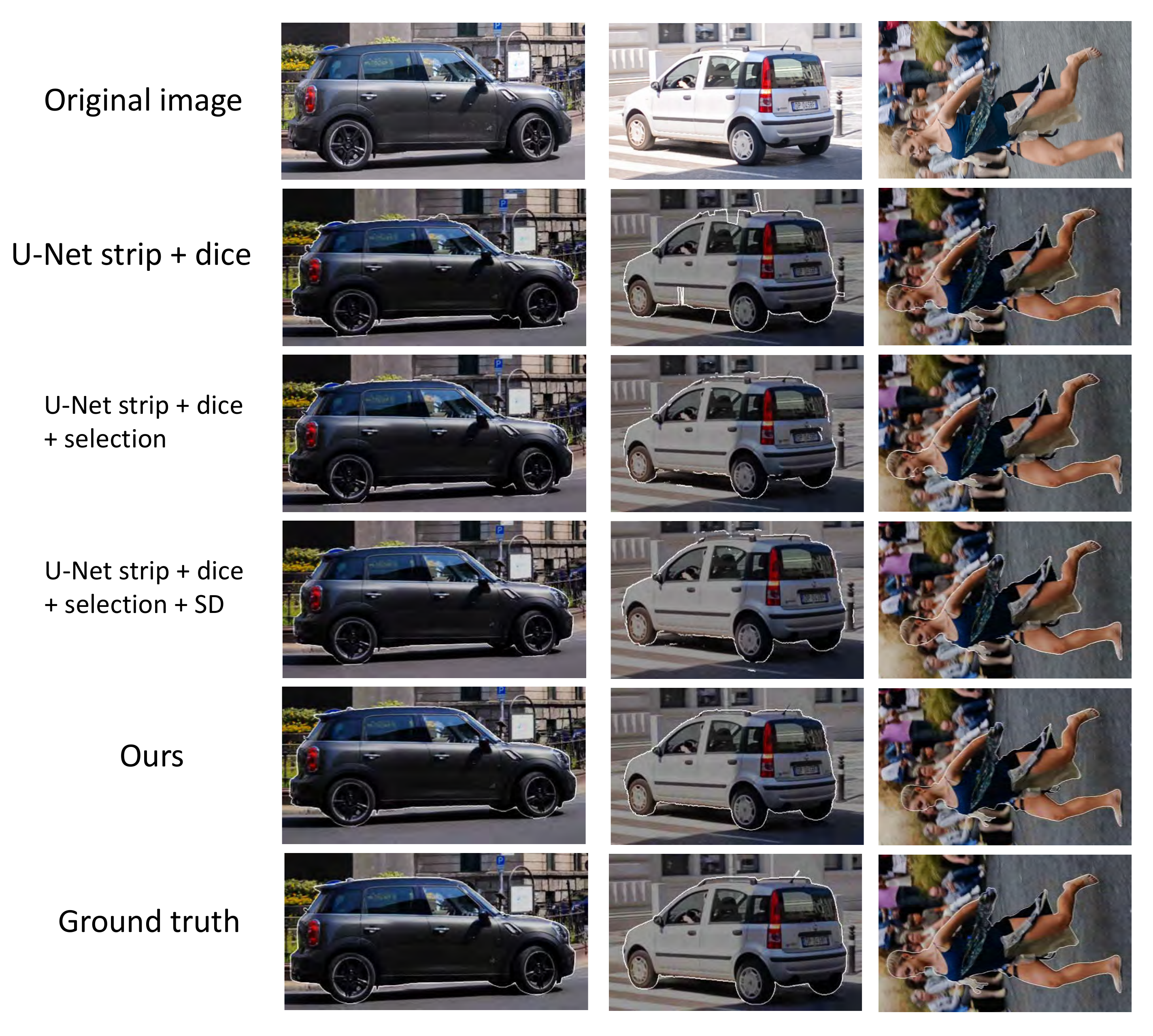}
\end{center}
\caption{Visualization of loss components. Rows from top to bottom are original images, U-Net strip $+$ dice which use weighted $l1$ and dice loss, U-Net strip $+$ dice $+$ selection, U-Net strip $+$ dice $+$ selection $+$ SD, Ours and Ground truth.}
\label{fig:abl}
\end{figure*}

Figure~\ref{fig:abl} compares the results with different losses. With only weighted $l1$ and dice loss, spurious boundaries are not suppressed and thus false positive exists. With the introduction of selection layer, the network select target boundaries from all potential ones so that spurious boundaries get ignored. Additionally, a closer prediction is observed with boundary distance loss. Lastly, with the introduction of C0 continuity and matching loss, a better result is obtained.

\subsection{Additional Qualitative results}
\begin{figure}[t]
\begin{center}
\vspace{-5pt}
   \includegraphics[width=1\linewidth]{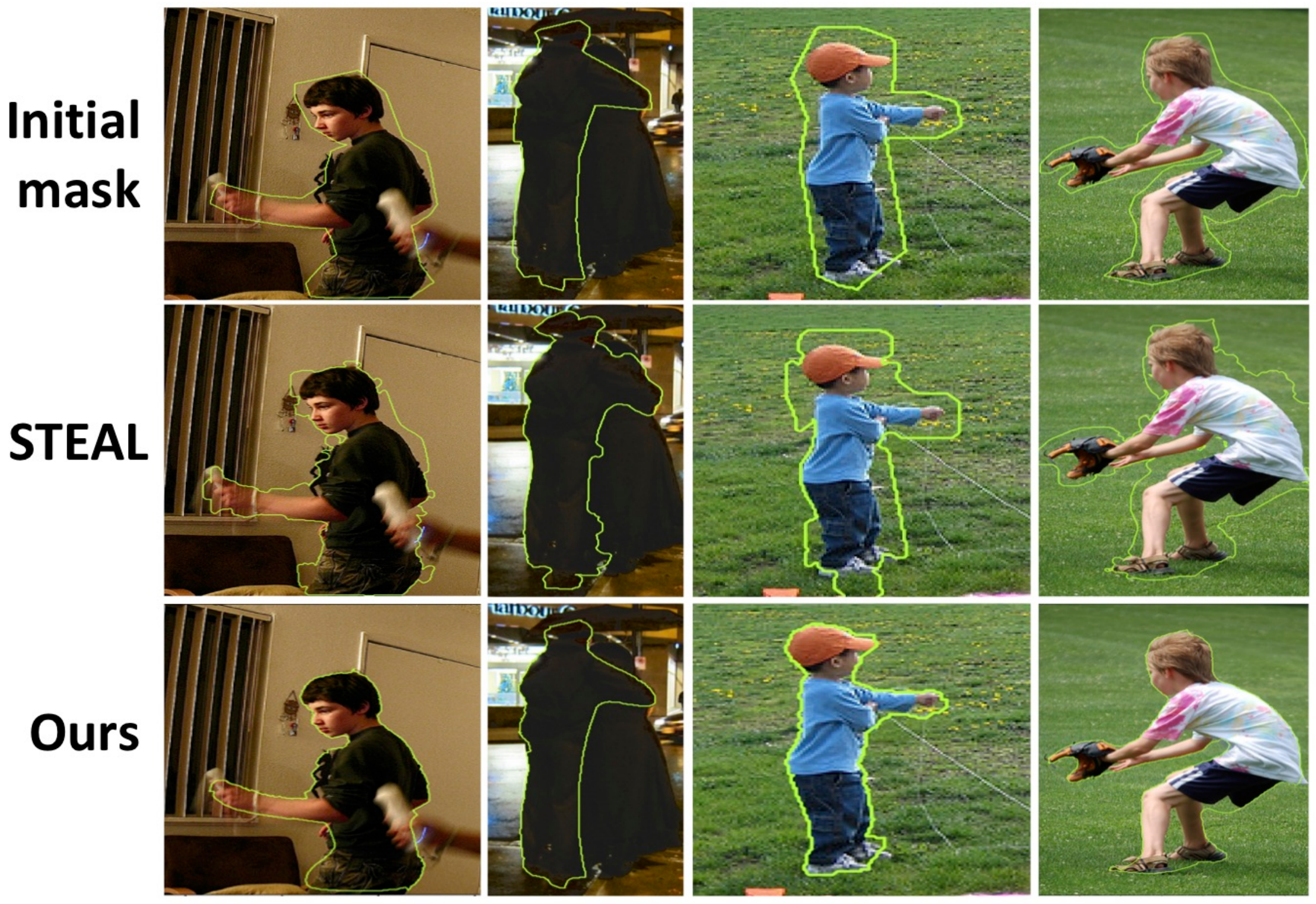}
\end{center}
\vspace{-5pt}
   \caption{Less accurate examples from COCO.}
\label{fig:coco}
\end{figure}

\begin{figure*}[t]

\begin{center}
\includegraphics[width=\linewidth]{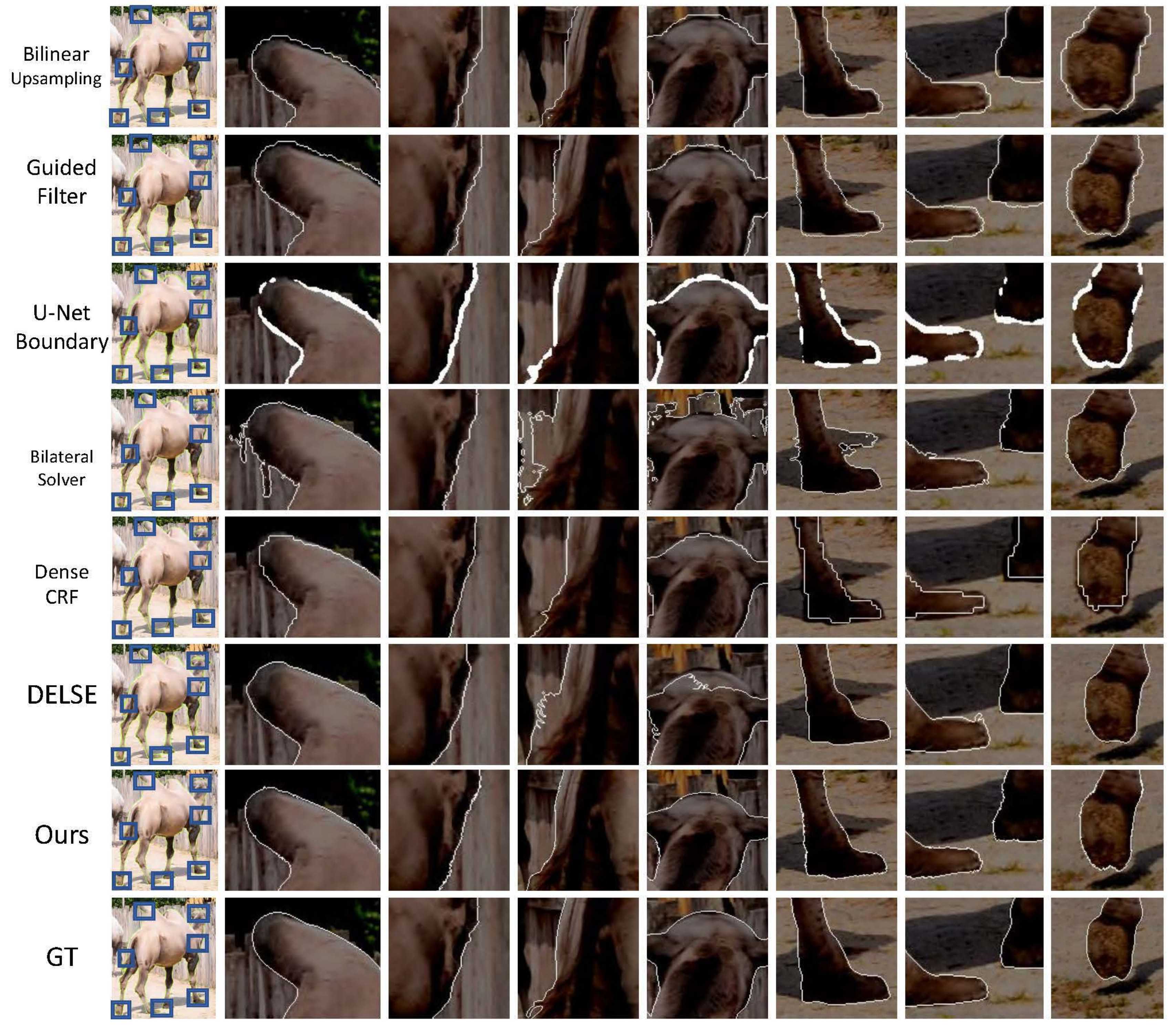}
\end{center}
\caption{Multi-region visualization on DAVIS 2016. The first column shows the whole image and the rest columns are the enlarged box regions. Boundaries are highlighted in white. Notice that our approach has closer prediction than methods like bilinear upsampling and guided filter, has less spurious boundaries than DELSE and bilateral solver, and thinner boundaries than U-Net Boundary.}
\label{fig:davis_whole}
\end{figure*}

\begin{figure*}[t]

\begin{center}
\includegraphics[width=\linewidth]{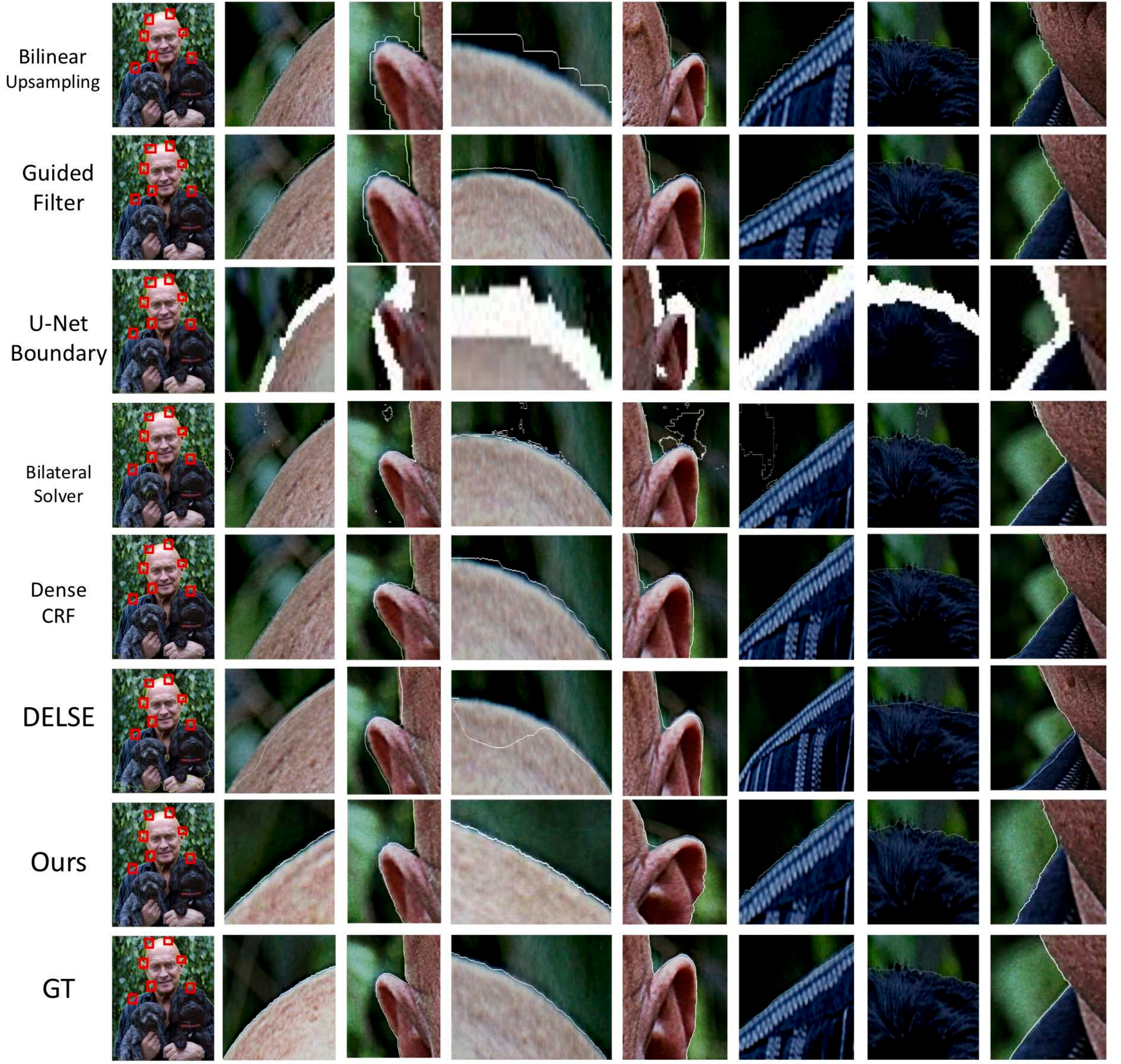}
\end{center}
\caption{Multi-region visualization on PixaHR. The first column shows the whole image and the rest columns are the enlarged box regions. Boundaries are highlighted in white. Notice that our approach has closer prediction than methods like bilinear upsampling and guided filter, has less spurious boundaries than DELSE and bilateral solver, and thinner boundaries than U-Net Boundary.}
\label{fig:pixa_whole}
\end{figure*}

\begin{figure*}[t]

\begin{center}
\includegraphics[width=\linewidth]{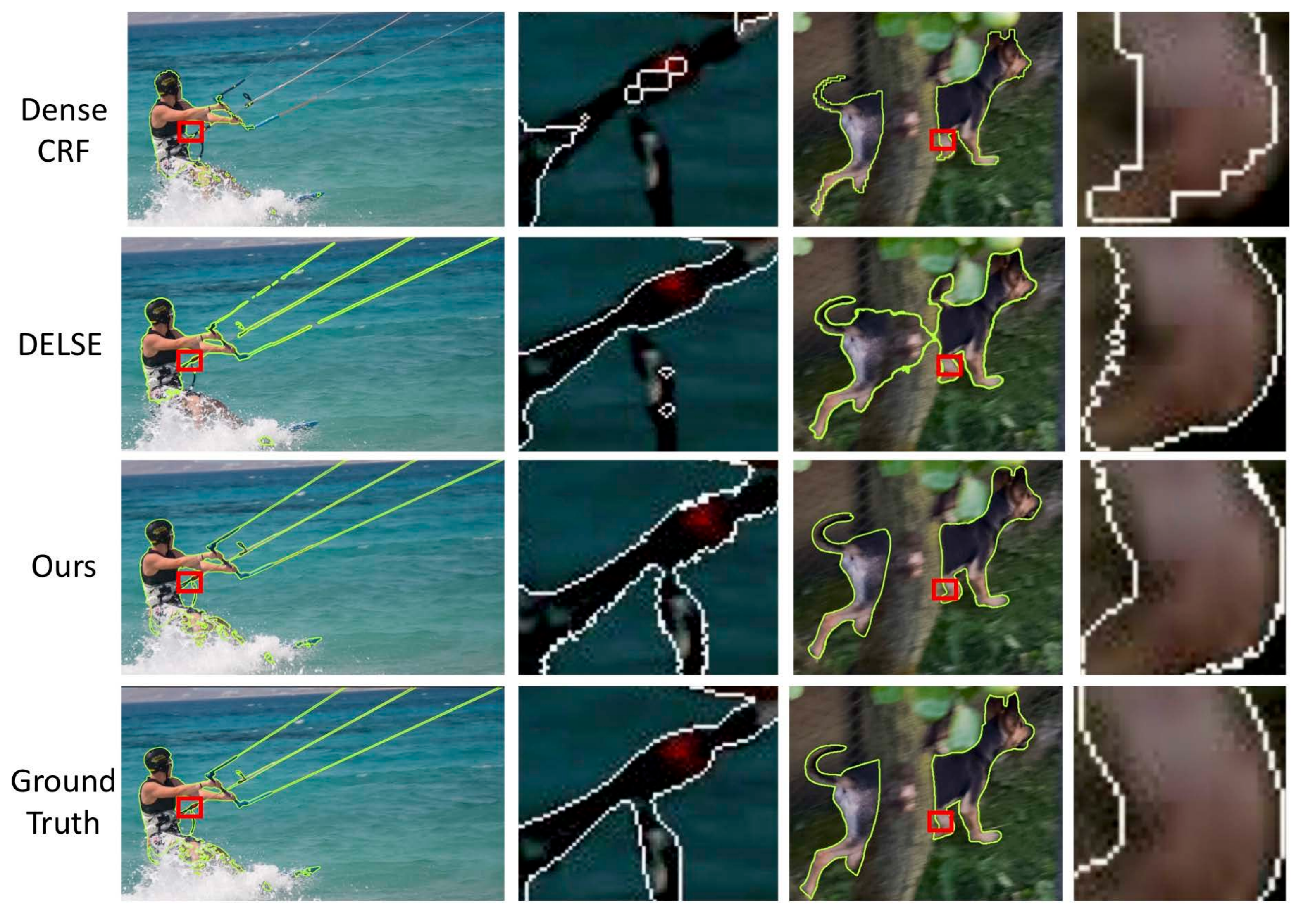}
\end{center}
\caption{Additional visualization on DAVIS 2016. We first show the whole boundary visualization and then show the enlarged box region. The boundaries in the enlarged regions are displayed in white. Notice that for complicated topology, our approach still has better result than the baselines.}
\label{fig:davis}
\end{figure*}

\begin{figure*}[t]

\begin{center}
\includegraphics[width=\linewidth]{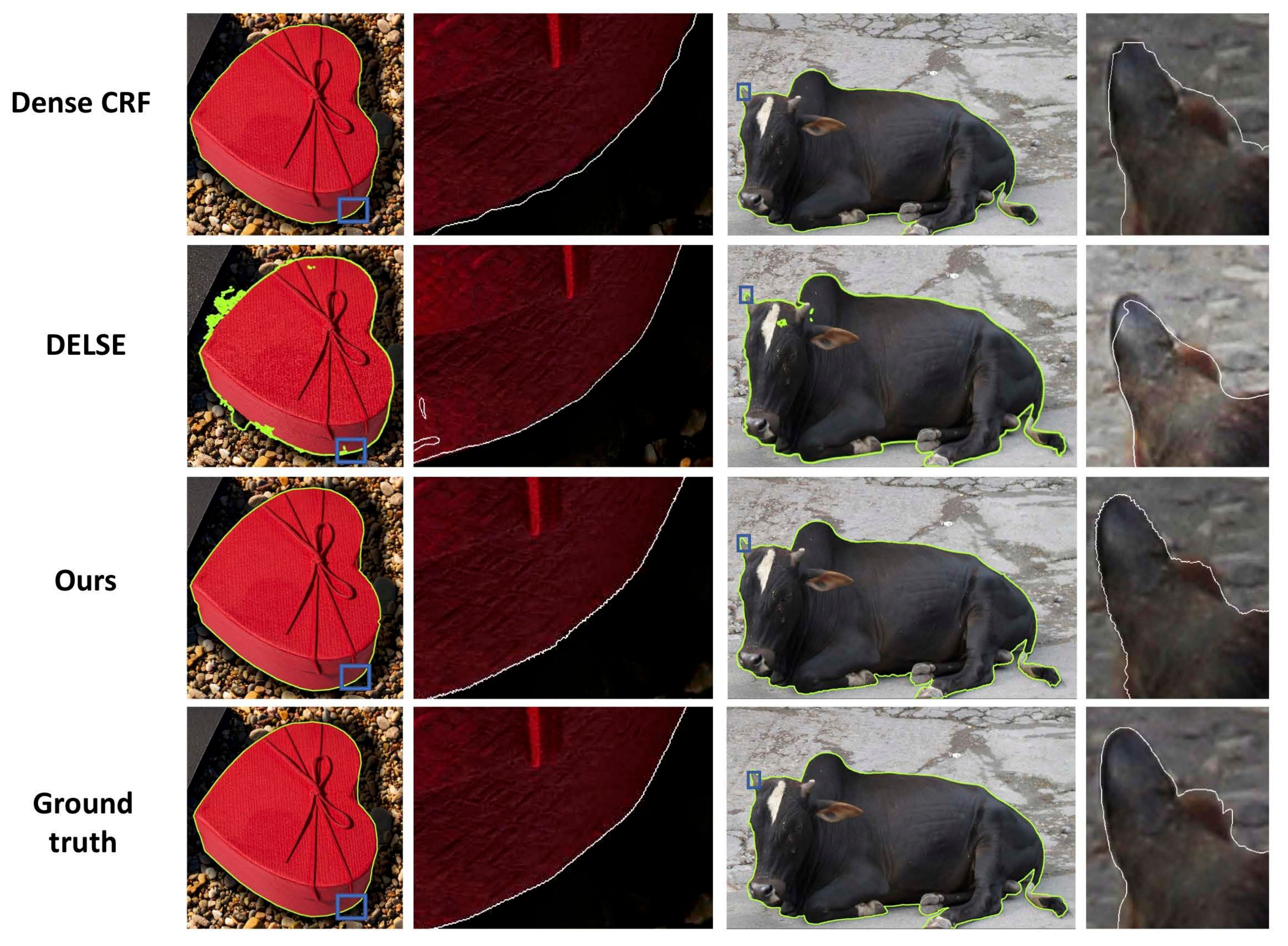}
\end{center}
\caption{Additional visualization on PixaHR $32 \times$. We first show the whole boundary visualization and then show the enlarged box region. The boundaries in the enlarged regions are displayed in white. Notice that our approach makes smoother prediction than dense CRF and less false positive than DELSE.}
\label{fig:pixa1}
\end{figure*}

\begin{figure*}[t]
\begin{center}
\includegraphics[width=\linewidth]{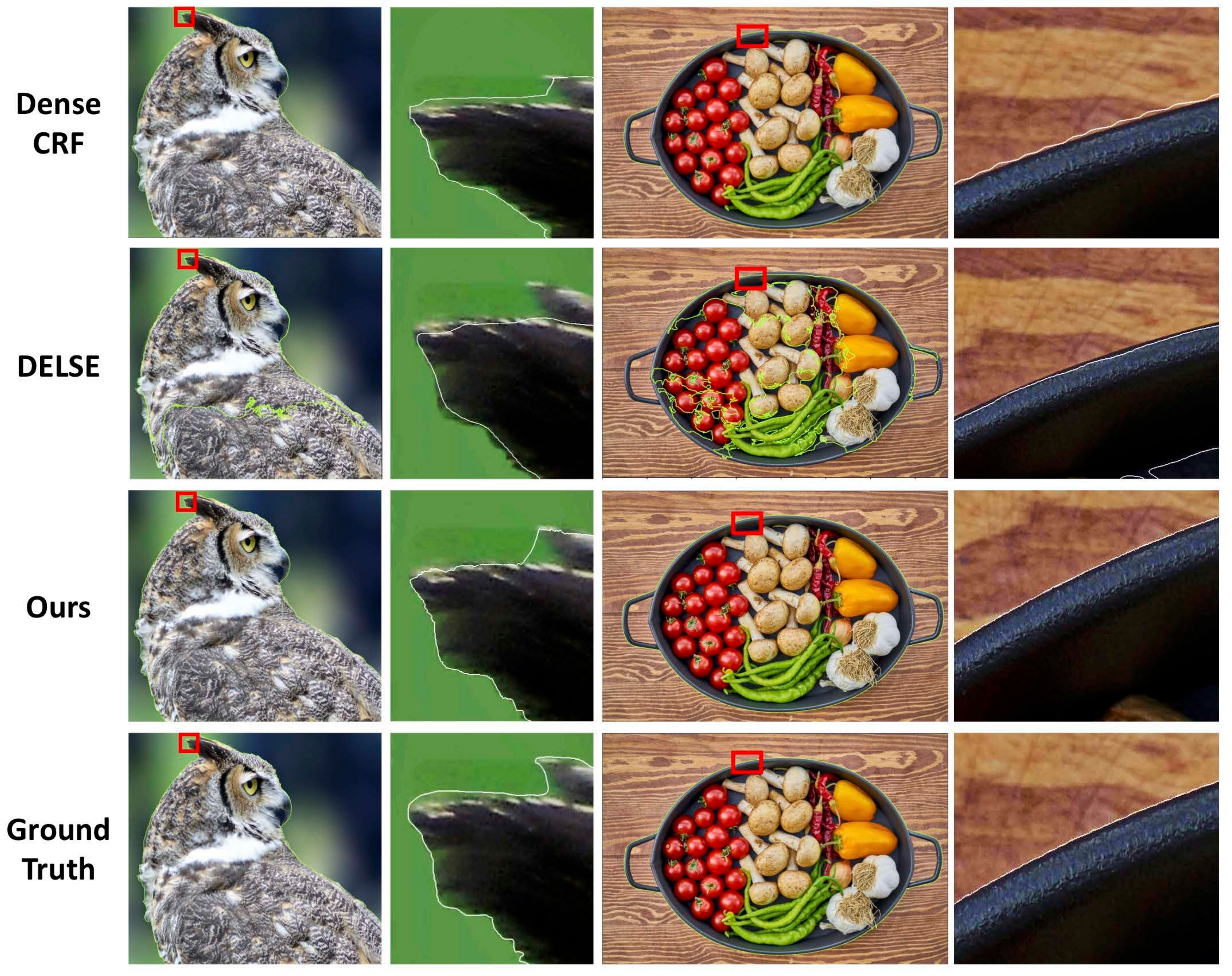}
\end{center}
\caption{Additional visualization on PixaHR $16 \times$. We first show the whole boundary visualization and then show the enlarged box region. The boundaries in the enlarged regions are displayed in white. Notice that our approach makes smoother prediction than dense CRF and less false positive than DELSE.}
\label{fig:pixa2}
\end{figure*}

Figure~\ref{fig:davis_whole} and Figure~\ref{fig:pixa_whole} show the results among baselines in multiple regions. It is clear that our method achieves more accurate results than the baselines. In particular, our approach have smoother boundaries than U-Net boundary and less false positive than DELSE and bilateral solver. More visualization examples are displayed in Figure~\ref{fig:davis}, Figure~\ref{fig:pixa1} and Figure~\ref{fig:pixa2}. Less accurate initial mask results on COCO is shown in Figure~\ref{fig:coco}.

\end{document}